\documentclass{sig-alternate-05-2015}
\usepackage{flushend}
\makeatletter
\def\@copyrightspace{\relax}
\makeatother
\begin{document}
\title{Compensating for Large In-Plane Rotations in Natural Images}
\numberofauthors{3} 
\author{
\alignauthor Lokesh Boominathan\\
       \affaddr{Video Analytics Lab}\\
       \affaddr{Indian Institute of Science}\\
       \affaddr{Bangalore, INDIA - 560012}\\
       \affaddr{boominathanlokesh@gmail.com}
\alignauthor Suraj Srinivas\\
       \affaddr{Video Analytics Lab}\\
       \affaddr{Indian Institute of Science}\\
       \affaddr{Bangalore, INDIA - 560012}\\
       \affaddr{surajsrinivas@grads.cds.iisc.ac.in}
\alignauthor R. Venkatesh Babu\\
       \affaddr{Video Analytics Lab}\\
       \affaddr{Indian Institute of Science}\\
       \affaddr{Bangalore, INDIA - 560012}\\
       \affaddr{venky@cds.iisc.ac.in}
}
\maketitle
\begin{abstract}
 Rotation invariance has been studied in the computer vision community primarily in the context of small in-plane rotations. This is usually achieved by building invariant image features. However, the problem of achieving invariance for large rotation angles remains largely unexplored. In this work, we tackle this problem by directly \textit{compensating} for large rotations, as opposed to building invariant features. This is inspired by the neuro-scientific concept of \textit{mental rotation}, which humans use to compare pairs of rotated objects. Our contributions here are three-fold. First, we train a Convolutional Neural Network (CNN) to detect image rotations. We find that generic CNN architectures are not suitable for this purpose. To this end, we introduce a convolutional template layer, which learns representations for canonical `unrotated' images. Second, we use Bayesian Optimization to quickly sift through a large number of candidate images to find the canonical `unrotated' image. Third, we use this method to achieve robustness to large angles in an image retrieval scenario. Our method is task-agnostic, and can be used as a pre-processing step in any computer vision system.
\end{abstract}
\section{Introduction}
Computer vision tasks are generally considered hard due to the presence of various nuisance factors such as pose, illumination, rotation, translation, among many others. Recently, very large and deep convolutional neural networks have achieved state of the art performance in visual recognition tasks \cite{krizhevsky2012imagenet}. However, it has been found that these networks are not completely invariant to factors like in-plane rotation \cite{gong2014multi}, simply because carefully curated images like those in the ILSVRC dataset do not often contain such large variations in angle \cite{ILSVRC15}. Even augmenting the training set with rotated images would be largely fruitless as the test / validation sets would not contain such rotated images. However, real-world photographs such as those on the web, or those taken by a user on a smartphone (or a first-person camera like GoPro) can naturally have large in-plane rotations. How do we make classifiers invariant to such large rotations?

\begin{figure}[t]
	\centering
	\includegraphics[width = 7cm]{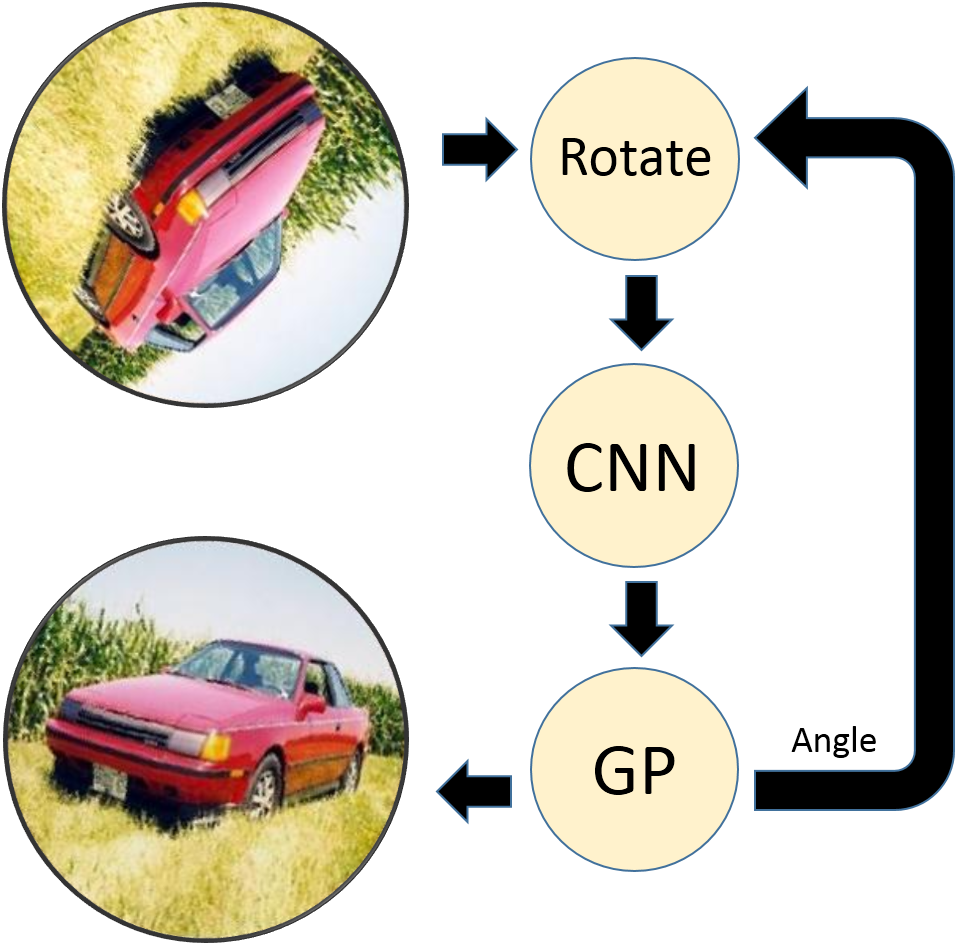}
	\caption{{Illustration of the overall method. Our method uses a CNN trained to detect rotations along with a GP-based Bayesian Optimizer to compute the correct angle of rotation.}}
	\label{fig:method}
\end{figure}

The traditional way of dealing with rotations involves designing the primary computer vision system to be rotation invariant. Image features such as SIFT \cite{lowe2004distinctive} and HoG \cite{dalal2005histograms} are designed to achieve invariance to small rotation angles. However, this becomes more and more difficult to design as the angle of rotations increase. Here we consider an alternate method, where we directly compensate for the rotation in the given image. The `corrected' image can then be fed into the primary computer vision system to deal with the task at hand. Hence, rotation compensation can be seen as a pre-processing step.

There have been very few recent works which explicitly tackle rotation invariance as a problem in itself. Jaderberg \textit{et al.} \cite{jaderberg2015spatial} proposed Spatial Transformer Networks, where they introduce a layer which implicitly learns rotation and scale invariance in a CNN-based framework. However, as was remarked earlier, this would only be effective for small angles if the database has only small rotations. Ding and Taylor \cite{ding2014mental} previously used the concept of mental rotations to estimate the relative rotation between two images. Our work, on the other hand, tries to estimate the absolute rotation of a single image. Chandrasekhar \textit{et al.} \cite{chandrasekhar2015practical} developed methods to compensate for large rotations in case of image retrieval. It involved creating multiple copies of the same image and then pooling their representations together. Our approach, on the other hand, can work for all tasks, including retrieval, classification and object detection. Sohn and Lee \cite{sohn2012learning} proposed to incorporate linear transformations into an RBM-based framework, in order to learn features invariant to those transformations. However, their method focuses only on small rotations (or other small local transformations which can be approximated as linear transformations) and is not well-suited for large rotations. Furthermore, \cite{sohn2012learning} achieves invariance only for the predefined transformations and not for any angle in general. 

There have also been a few works which deal with large in-plane rotations. Vailaya \textit{et al.} \cite{vailaya2002automatic} developed a method to automatically detect the underlying rotation angle of an image, but the method is restricted only to input images with four possible rotations that are multiples of $90^\circ$ ($0^\circ$,$90^\circ$,$180^\circ$ and $270^\circ$). Rotation detection algorithm proposed by Nagashima \textit{et al.} \cite{nagashima2007high} finds the relative rotation between two images, whereas our method can be used to detect the absolute rotation angle of a given input image. Methods such as  \cite{gong2014multi} and \cite{ahonen2009rotation} have been developed to get features that are robust to small rotations. 

\begin{figure}[t]
	\centering
	\includegraphics[width = 8cm,trim={0cm 0cm 0cm 0cm},clip]{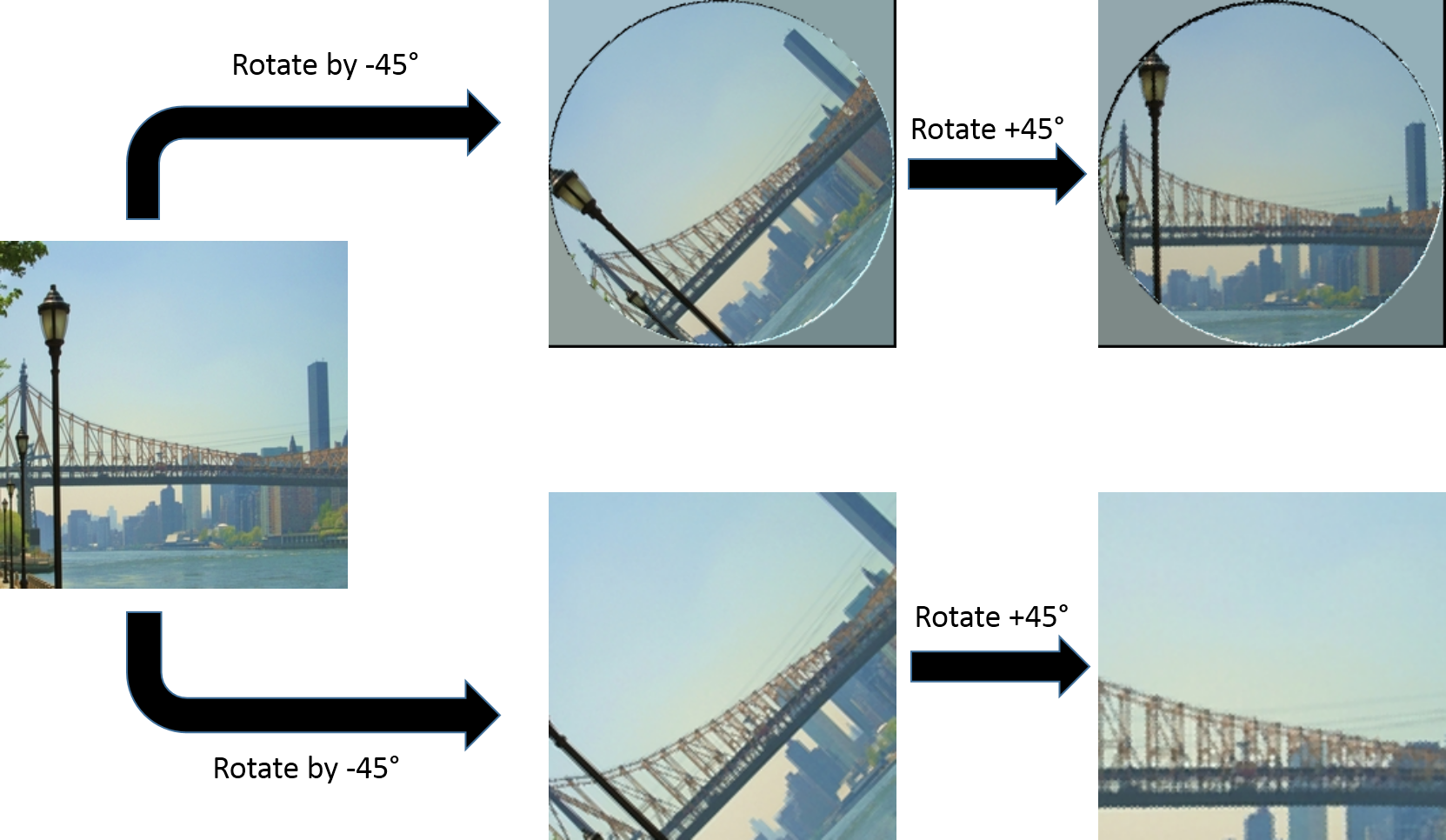}
	\caption{{Illustration of the effects of repeated rotation. We observe that rotating a circular patch in the image better preserves the image content, while the usual way of doing rotation has a zooming effect due to the implicit cropping.}}
	\label{fig:rotate}
\end{figure}

Our work is primarily inspired from the neuro-scientific concept of mental rotations. The neuroscience community has extensively studied the representations of spatial relationships in our brain, one of them being mental rotations. Shepard and Metzler \cite{shepard1971mental} first studied this phenomenon for 3D objects. Their study found that the time a subject takes to determine whether two objects were rotated versions of each other (or not) was proportional to the relative rotation between them. Shepard and Cooper \cite{shepard1986mental} further showed that this holds true for all kinds of rotations, including in-plane rotations. While these studies were done for pairs of objects, it is interesting to note that humans also have the capacity to estimate rotation of individual images of natural scenes. When presented with an image of a beach rotated by $90^\circ$, we can still identify the beach and observe that it is rotated by $90^\circ$. This leads us to believe that we \textit{may} also perform mental rotation on individual images, along with image pairs. While it is unclear whether we indeed use mental rotations for individual images, for this work we nonetheless use the same for building a system invariant to large rotations. 

One hypothesis for the case of individual images would be that we form mental representations of natural images and use those representations to determine whether an individual image is rotated or not. While it is unclear whether this is true for the human visual system, we nonetheless proceed to apply this principle to our case of computer vision.

In this paper, we shall use two terminologies extensively. The term \textbf{absolute zero} image would refer to the image which is in it's canonical representation and looks ``unrotated'' and ``straight''. The term \textbf{absolute angle} of an image would refer to the rotation angle of a given image relative to the \textit{absolute zero} image. Our task can now be stated as follows. Given an image $x(\theta)$, where $\theta$ is the unknown absolute angle, we wish to find $\theta$ and hence obtain $x(0)$, the \textit{absolute zero} image.

Our paper is organized as follows. Section 2 discusses a method to rotate images. Section 3 discusses performance of CNN features for varying angles of rotation. In Section 4, we describe our method of training a CNN to detect whether a given image is rotated or not. The Bayesian Optimization procedure is described in Section 5. Experiments and concluding remarks follow.

\section{How to rotate images?}
While working with images and their rotations, it becomes important to think about how we want to rotate images. The usual way of performing image rotation can change image dimensions depending on the amount of rotation. For example, a $0^\circ$ rotated image will always be larger than a $45^\circ$ image. As a result, when both are scaled to the same dimensions, the $45^\circ$ rotated image would appear to be zoomed in. Further, let $x(\theta)$ be an image rotated by angle $\theta$. We can see by Figure \ref{fig:rotate} that we cannot recover $x(0)$ by first rotating an image by $+45^\circ$ then rotating back by $-45^\circ$. This is because $x(45)$ is less informative than $x(0)$ due to the implicit cropping effect.

We can minimize the amount of cropping done by considering only square images - these can easily be obtained by the largest central square crop in the image. However, to completely solve the issue, we consider the central circular region in each image for rotation. Of course, this means that the 4 corners of the image would contain no information, as shown in Figure \ref{fig:rotate}. However, this makes the rotation operator better behaved, which is what we need when working with large rotations. Once our system has estimated the absolute angle of the image, we can use the usual way of doing rotations to compensate for the rotation to feed it to the next stage.

\section{Do CNN features work well for rotated images?}
Given that classifiers are trained on images at a particular canonical angle, they are not expected to work well for other images. While this has been established for classification by \cite{gong2014multi}, we wish to investigate whether this holds true for the learnt features in a deep network in general. In other words, do the deep learnt features capture relevant semantics only when the image is in it's canonical angle? To test this, we consider an image retrieval scenario. We use CNN features to perform retrieval, as shown in \cite{babenko2014neural}. 

We do two experiments. For a given dataset and a query image, we first rotate the query image through different angles and look at the retrieved list of images. Second, we rotate both the query image and \textbf{all} the dataset images so that there is \textbf{no relative rotation} between them.

\begin{figure}[t]
\centering
\includegraphics[width = 7cm,trim={0cm 0cm 0cm 0cm},clip]{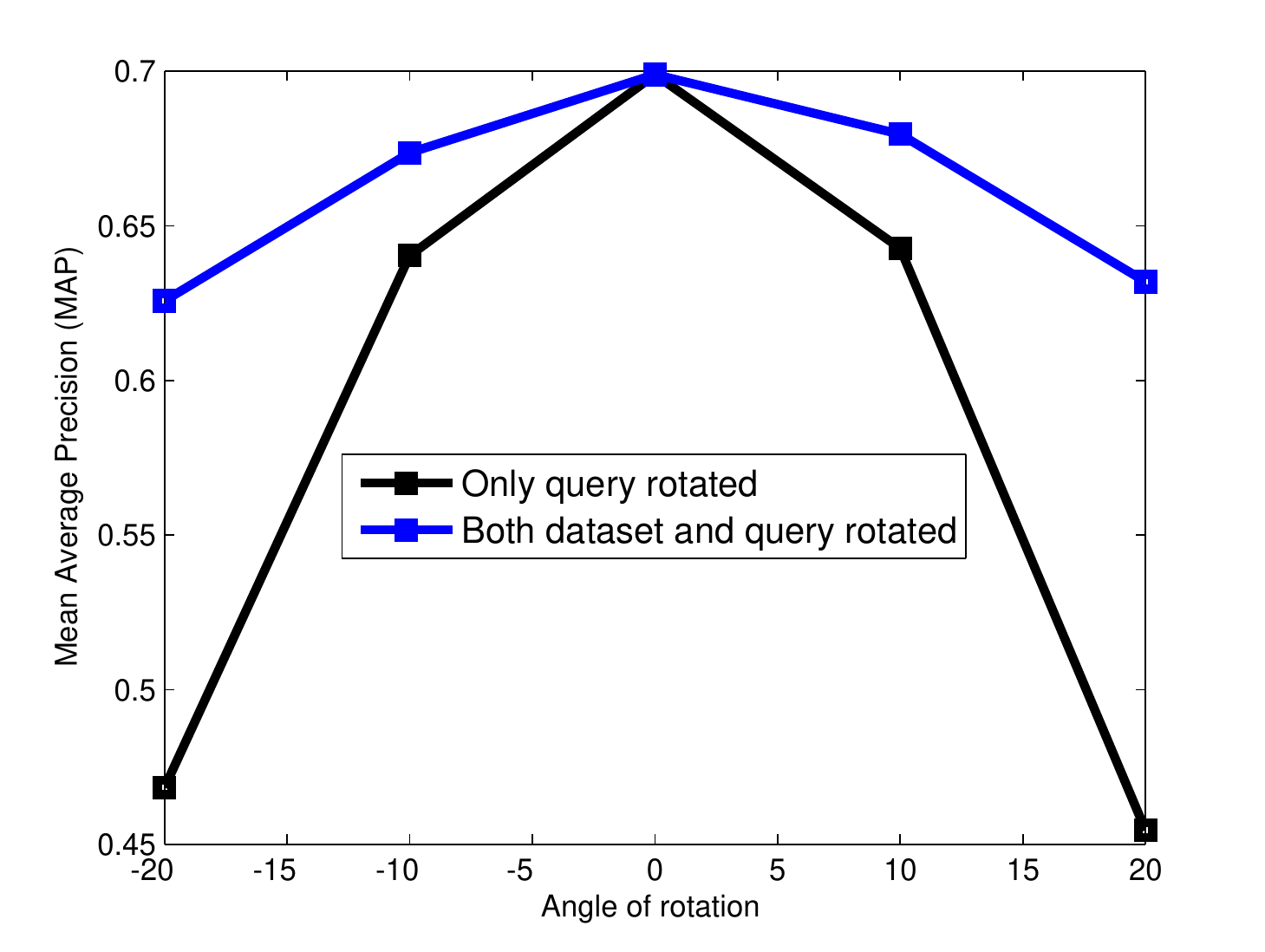}
\caption{Plot of MAP scores of image retrieval vs angle of rotation. While both curves fall with increasing angle, we see that it is smaller when there is no relative rotation between dataset and query.}
\label{fig:rot}
\end{figure}

We perform this experiment on the Holidays dataset and the fc7 layer of the AlexNet CNN \cite{krizhevsky2012imagenet}. As shown in Figure \ref{fig:rot}, both curves fall with increasing angle. However, the black curve has a steeper fall because of the mis-match between angles in the query image and dataset. The blue curve, on the other hand, has a slower fall even when there is no relative rotation between the query and dataset. This experiment shows that CNN features do not perform well for images which are not in their canonical form. This implies that higher level CNN features do not code for rotation, which in turn means that one cannot use them to detect rotation.

\section{Rotation detection}
Rotation compensation of an image can be achieved in one of two ways. We can either
\begin{itemize}
\item train a model to \emph{estimate} rotation of a given image. Use that estimate to obtain the absolute-zero image.
\item train a model to \emph{detect} whether a given image is rotated or not. Given an image, rotate that image through all possible angles. Select the image which the model believes is \emph{not} rotated. 
\end{itemize}

Admittedly, the second option is more convoluted than the first. However, we still use this approach because it is simpler problem to solve. A rotation \emph{estimation} model has to learn internal representations of how images at \emph{all} absolute angles look like. A rotation \emph{detection} model just needs to have a representation of how an absolute-zero image looks like. Given that there can be large amount of variability in how an absolute-zero image itself may look like (varying image detail and nuisance factors), the task of rotation \emph{detection} itself is a challenging one to solve.

\begin{figure*}
	\centering
	\includegraphics[width = 18cm,trim={0cm 10cm 0cm 1.77cm},clip]{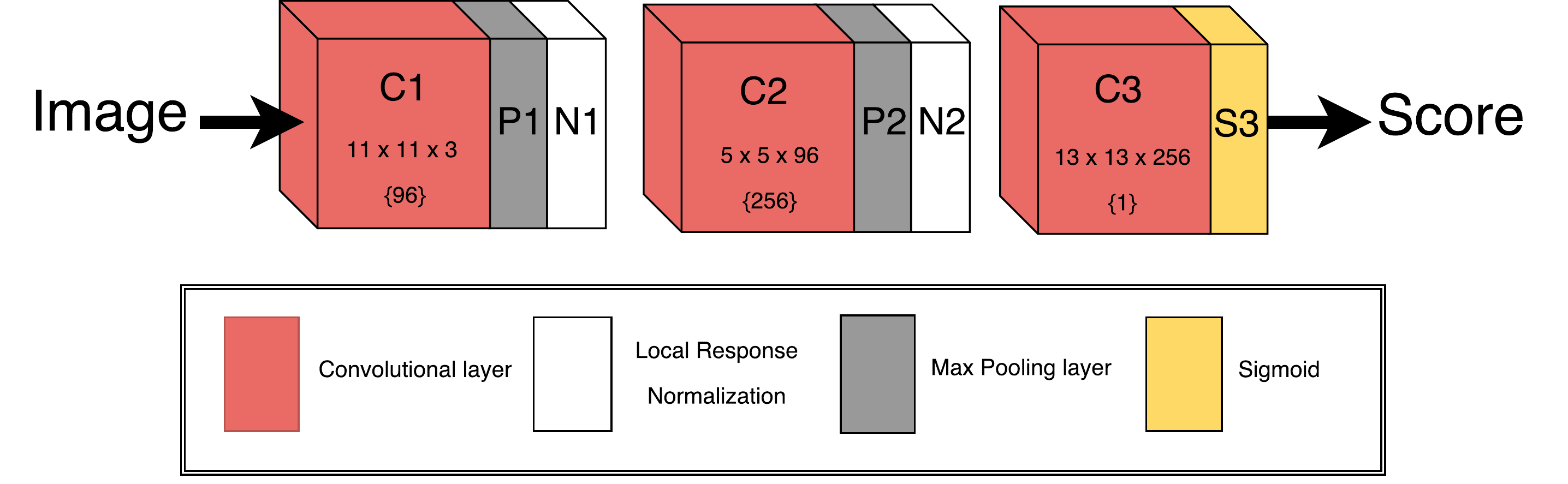}
	\caption{{The proposed CNN architecture for learning rotation detection. Here, layer C3 denotes the \textit{template} layer.}}
	\label{fig:net}
\end{figure*}

\subsection{CNN for rotation detection}
What does it take for a system to detect whether a given image is rotated or not? If we consider outdoor images, we can consider some simple rules that might work well. For example, if the top half of the image is blue, and the bottom part is green, it is possible that the underlying image is an absolute zero image (of a park, maybe). Our model leverages this observations to learn such \textbf{image templates} in a CNN framework. Unlike regular convolutional layers which have small receptive fields, the template layer looks at the entire image at once. This contains the internal representation that the CNN learns about absolute-zero images. Many such templates can account for the variability in image content.

As a concrete example, consider the network in Figure \ref{fig:net}. If a $224\times224$ RGB image if fed to this CNN, the feature size after the second convolutional layer is $13\times13\times256$. This is exactly the size of the template layer (third convolutional layer). The output of that layer is a scalar, which is the network's final output.

The popular approach for producing scalar outputs in CNNs is to use a fully connected layer. This is a vectorized representation which does not necessarily contain any spatial information about the scene. Given that in-plane rotation of an image relies heavily on spatial orientation, we use the proposed template layer to enforce a 2D representation in the CNN. Further, fully connected layers introduce a large number of parameters in the model. This stifles learning and causes our model to overfit. Template layers, on the other hand, have a much smaller number of parameters which enables good generalization. 

\subsection{Generalization behaviour}

Our architecture (Figure \ref{fig:net}) consists of three convolutional layers. While the first two layers have weights identical to that of AlexNet \cite{krizhevsky2012imagenet}, the third layer is a convolutional \textit{template} layer whose output is a scalar. We apply a sigmoid non-linearity on that scalar so that the overall output of the network lies between 0 and 1. We call this output as the \emph{score} for a given image. A score of \verb+0+ would correspond to absolute zero images, while higher scores map to images with an underlying rotation. We use the cross-entropy loss function to train the objective. 

\subsection{Dataset for rotation detection}

Given that we train a CNN to perform rotation \textit{detection}, we construct the training data as follows. We take 5000 random images from the MIT Places dataset \cite{zhou2014learning} and perform 5 \textbf{random} rotations (from $-180^\circ$ to $180^\circ$ ) according to the technique described in the section on how to rotate images. This increases the size of training data to $5000 \times 6 = 30000$ images. Assuming that the images sampled from Places dataset are all aligned at an absolute angle of zero, these can be given a label of \verb+0+, whereas all the rotated images can be given a label of \verb+1+. This makes it a simple \textbf{binary classification} problem. The resulting trained CNN is termed as `3-CNN'(\verb+3+ indicates the number of layers). 

However, we note that not all images in the dataset are at an absolute angle of zero. There may exist images with small rotations. To compensate for this, we assign all images with rotation between $-10^\circ$ and $+10^\circ$ with a label \verb+0+. The CNN trained on this dataset is termed as `3T-CNN'(\verb+3+ indicates the number of layers and \verb+T+ denotes the inclusion of angles between $-10^\circ$ and $+10^\circ$). 

\subsection{Generalization behaviour}
%\begin{figure}
%	\centering
%	\includegraphics[width = 9cm,trim={1cm 0cm 0cm 1cm},clip]{abs_new.pdf}
%	\vspace{-1cm}
%	\caption{{Plot of median CNN scores vs absolute angle of image. As expected, the global minimum is at an absolute angle of zero degrees. We also observe that training the CNN with $\pm 10^\circ$ images with label `0' makes the second curve smoother than the first, while pushing down that median at zero. We also plot a similar curve for the HoG + SVM classifier and see that it is less discriminative in general.}}
%	\label{fig:abs}
%\end{figure}

%	\centering
%	\includegraphics[width = 18cm,trim={2cm 0cm 1cm 0cm},clip]{hist_new.pdf}
%	\caption{{Normalized histogram of the absolute angle of corrected images. The perfect predictor is a delta distribution centered at 0 degrees. We see that the 6-layer CNN performs the best, closely followed by the 3-layers CNNs. We also test cross-dataset performance by evaluating on the Holidays dataset.}}
%	\label{fig:hist}
%\end{figure*}

\begin{figure}
	\centering
	\includegraphics[width = 9cm,trim={1cm 0cm 0cm 1cm},clip]{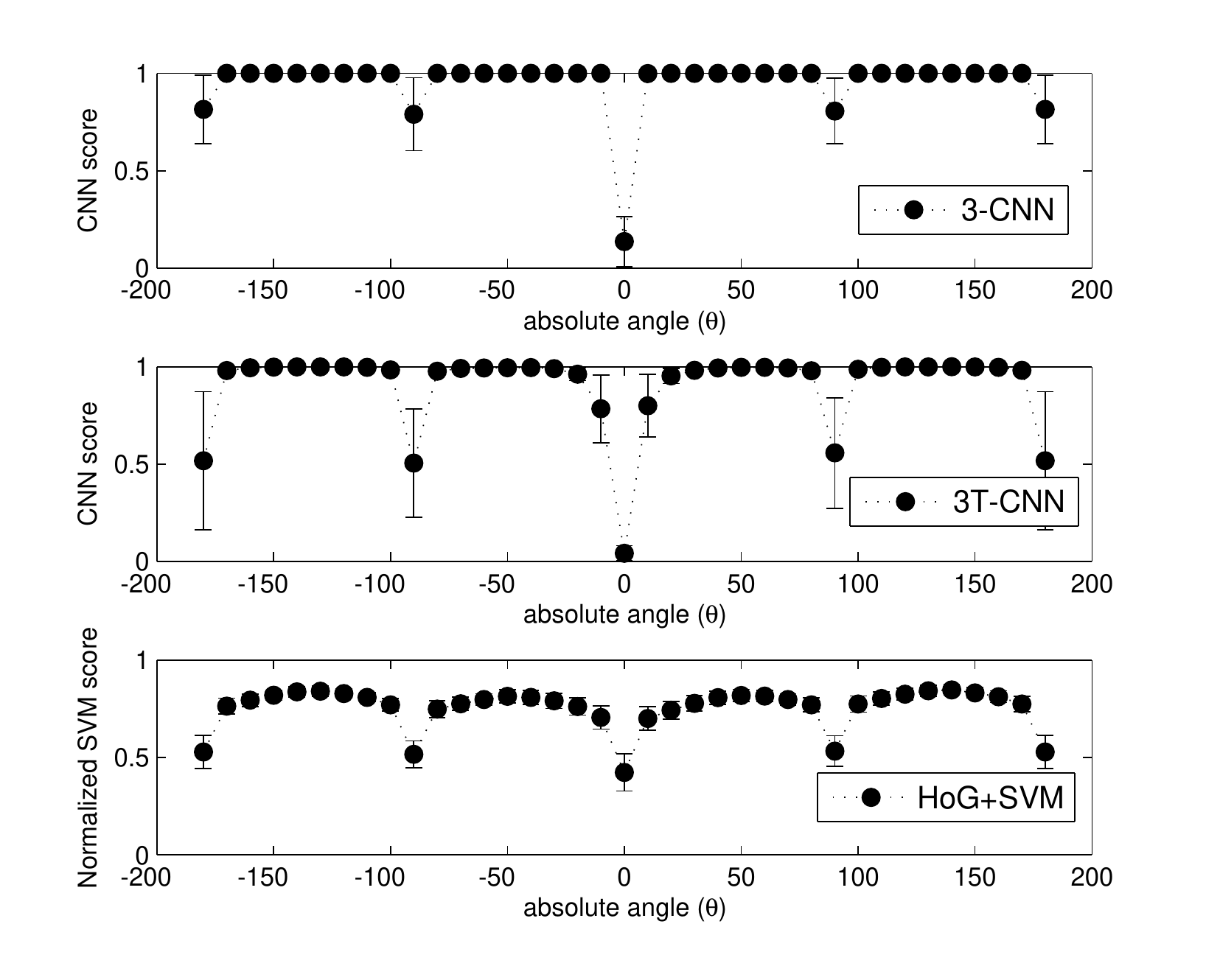}
	\vspace{-1cm}
	\caption{{Plot of median CNN scores vs absolute angle of image. As expected, the global minimum is at an absolute angle of zero degrees. We also observe that training the CNN with $\pm 10^\circ$ images with label `0' makes the second curve smoother than the first, while pushing down that median at zero. We also plot a similar curve for the HoG + SVM classifier and see that it is less discriminative in general.}}
	\label{fig:abs}
\end{figure}

To study the generalization behaviour of the CNNs considered above, we plot a curve of absolute angle against the output (or score) of the CNN over a validation dataset of 600 images. Each image in the set of $600$ is rotated from $[-180^\circ,170^\circ]$ range with a gap of $10^\circ$, making it $36$ angles in total. Hence the total dataset size is $600 \times 36 = 21600$ images. Figure \ref{fig:abs} shows a plot of CNN score against input image's absolute angle. We find that the first plot has a very steep fall around absolute zero angle. While this is what the classifier was trained for, we see that this makes searching for \textit{absolute zero} very difficult. For example, if $x(0)$ is the required output and we evaluate $x(10)$, and find that to be as high as $x(250)$ (for example), then the only strategy that we can use to find $x(0)$ is exhaustive search. However, if $x(10)$ could somehow tell us that the minimum is close, we could do something more intelligent.

\begin{figure*}[t]
	\centering
	\includegraphics[width = 18cm,trim={2cm 0cm 1cm 0cm},clip]{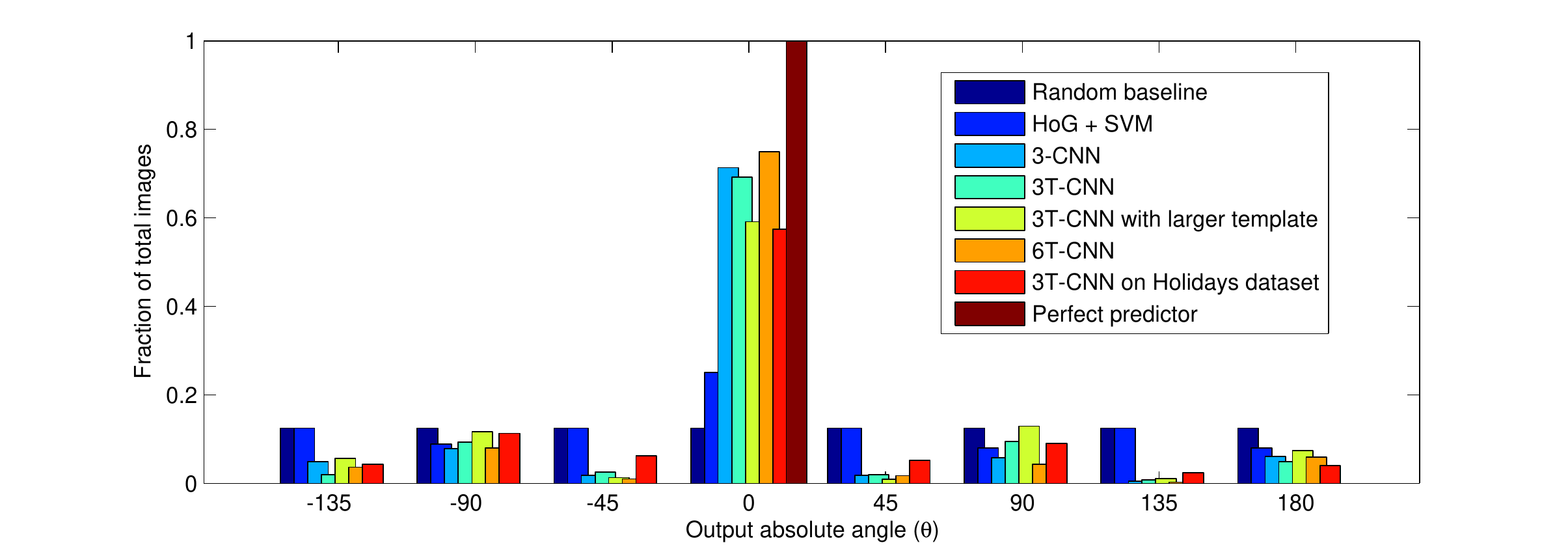}
	\caption{{Normalized histogram of the absolute angle of corrected images. The perfect predictor is a delta distribution centered at 0 degrees. We see that the 6-layer CNN performs the best, closely followed by the 3-layers CNNs. We also test cross-dataset performance by evaluating on the Holidays dataset.}}
	\label{fig:hist}
\end{figure*}

On the contrary, we find the second plot of Figure \ref{fig:abs} much better behaved. First, this makes the overall plot much smoother around \textit{absolute zero}, which helps in searching for the absolute-zero angle. This speed advantage is shown in Table \ref{tab:table}. Second, it shifts the median value at zero much lower than before, which ensures that an image at absolute zero almost always gets a very low value. However, we see that it is also more ``noisy'' as the median values for $\pm 90^\circ$ and $180^\circ$ are lower - which can cause mis-classification.

We also plot a similar curve for a simple baseline of HoG with an SVM classifier. The performance in general is worse for this classifier when compared to 3-CNN and 3T-CNN. This is indicated by the smaller gaps between the scores for zero and other angles.

\section{Searching for absolute-zero}
We can use the trained CNN discussed previously to directly obtain the \textit{absolute} image, by exhaustively searching from a pool of rotated images. Can we perform a more intelligent search instead?

Let function $f(\cdot)$ map an image to it's corresponding CNN score, and let $x(\theta)$ be the input image $x(\cdot)$ rotated by an angle $\theta$. We can formulate the search problem as the solution of the following equation.

\begin{equation}
	\theta_{min}=argmin_{\theta} ~f(x(\theta))
\end{equation}

Given that the function $f(x(\cdot))$ does not have an analytic expression, we treat it as a black-box function and use Bayesian Optimization (BO). BO is general framework for optimization of black-box functions. It is typically used for tasks like hyperparameter selection. 

Broadly speaking, BO can be thought of as a framework for intelligently sampling points in a closed interval, taking into account the uncertainty about the functional values. In this setting, we typically define an \textit{acquisition function} which balances between \textit{exploring} unknown regions with large uncertainty, and \textit{exploiting} regions close to the lowest found values. A Gaussian Process (GP) based regressor is used to approximate the black-box function's behaviour at unknown sample points. A practical guide to Bayesian Optimization can be found in \cite{snoek2012practical}.

\section{Experiments and Results}
We experimentally evaluate our method on the overall task of rotation compensation. We used the Caffe Deep learning framework \cite{jia2014caffe} for training the CNN and BayesOpt \cite{martinez2014bayesopt} for performing Bayesian Optimization. For the Bayesian Optimization, we divided the range of angles from $-180$ to $180$ into a discrete grid containing $180$ points. This gives us an effective resolution of $2^\circ$. The optimizer samples a maximum of $30$ angles, after which it returns an estimated global minimum. We also use a lower threshold on the function value so that the sampling may cease when a suitably low functional value is found soon.

\subsection{Performance Evaluation}

We evaluate the effectiveness of our rotation compensation method by using a test set of $2000$ random images from the Places dataset. Each image in the set was rotated to $[0,~\pm 45,~\pm 90,~\pm 135, 180]$ degrees, yielding $2000 \times 8 = 16000$ images overall. 

We plot our results as a histogram of the absolute angles of the compensated images, as in Figure \ref{fig:hist}. We compare performances of different variations discussed in Section $4$. We use as a baseline the classical HoG feature \cite{dalal2005histograms} with an SVM classifier, instead of using the CNN. Curiously, we see that HoG performs almost as well as the CNNs for $\pm 90^\circ$ and $180^\circ$, and as good as random at other angles. 

Additionally, it is also plausible that a larger template layer in the CNN may be able to perform better. To test the same, we compare against such a model by removing the max-pooling in layer 2. This increases the template size to $27 \times 27$, up from $13 \times 13$. However, it failed to perform as well as the smaller templates. We attribute this to the increased number of parameters which may make learning difficult. 

We also use a deeper CNN (6T-CNN) and see that it indeed performs the best. However, it is also considerably slower because of it's depth. Table \ref{tab:table} shows this result. We also observe that the 3T-CNN takes the least amount of samples to compensate for rotation as it's $f(\cdot)$ curve is smoother, making it easy to find the minimum.

\begin{table}[h]
	\centering
	\begin{tabular}{|c|c|c|}
		\hline
		\textbf{CNN Model} & \textbf{Total Time (sec)} & \textbf{No. of images}  \\ \hline
		3-CNN & 22.97 & 19.4\\ \hline
		3T-CNN & 14.44 & 9\\ \hline
		6T-CNN & 79.81 & 9.865\\ \hline
	\end{tabular}
	\caption{{A comparison between the various methods used for compensating rotation in an image. The first column shows the total time taken for the entire compensation process, and the second column shows the average number of images sampled by the Gaussian Process. As expected, the 3-layer CNNs are faster.}}
	\label{tab:table}
\end{table}

The performance of 3T-CNN was also evaluated on Holidays dataset \cite{jegou2008hamming} in order to account for generalization across datasets. As can be observed in Figure \ref{fig:hist}, the method performs adequately well, although not as well as it performed for Places-based dataset.

\subsection{Application to Image Retrieval}

\begin{figure}[h]
	%\centering
	\includegraphics[width = 9cm]{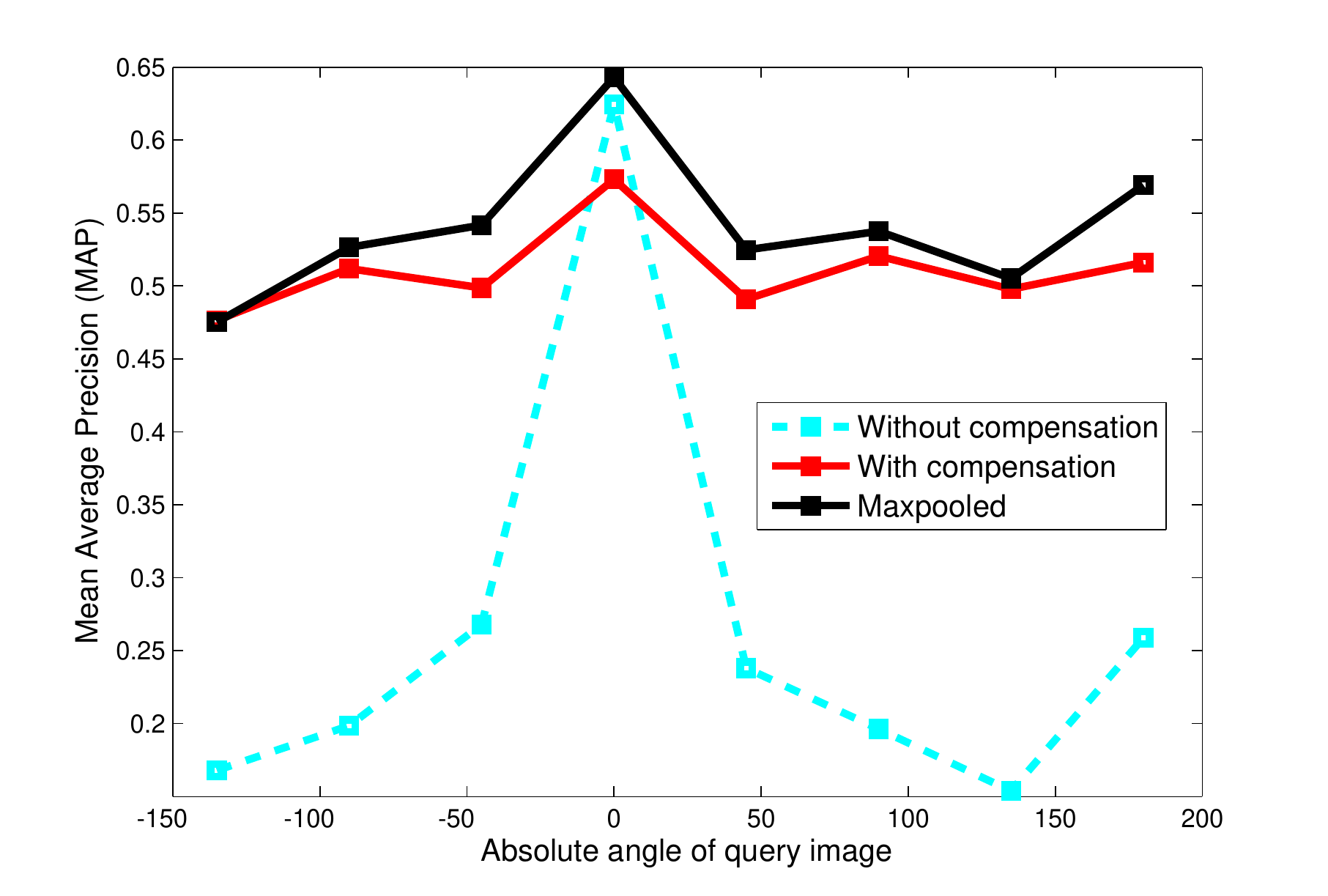}
	\caption{{Plot of retrieval performance with changing query image angle. We see that using rotation compensation can drastically improve performance for query images with large rotation.}}
	\label{fig:rot_inv}
\end{figure}

In this section, we apply our method to the task of image retrieval to bring about robustness (or invariance) to large rotations.

For this task, we use the Holidays dataset and perform retrieval using the \verb+fc7+ layer of AlexNet, in a manner similar to \cite{babenko2014neural}. To test for robustness to rotation, the query image is rotated through different angles and the Mean Average Precision (MAP) values are calculated. As shown in Figure \ref{fig:rot_inv}, the blue curve falls with increasing angle, thus exhibiting variance to rotation. The MAP scores for images corrected using the 3T-CNN (red curve) shows higher values at $~\pm 90^\circ$ and $180^\circ$, but suffers a small fall at other angles. Chandrasekhar \textit{et al.} \cite{chandrasekhar2015practical} had shown that max-pooling \verb+fc7+ features of multiple rotated images helps in increasing robustness to rotation. We use a similar approach here and maxpool the \verb+fc7+ feature of the query image with it's rotation compensated version. As can been observed in Figure \ref{fig:rot_inv}, the max-pooled features show more robustness to rotation as compared to both the un-rotated set of images as well as the corrected set of images. It is also important to note that the rotated images of Holidays dataset were compensated using a CNN that was trained with a subset of Places dataset. This shows that the method can be generalized across datasets to obtain rotation robustness to large rotations.

\subsection{Behaviour at $\pm90^\circ$ and $180^\circ$}
Throughout our empirical evaluation, in Figure \ref{fig:abs}, \ref{fig:hist} and \ref{fig:rot_inv}, we noted the strange behaviour at $\pm90^\circ$ and $180^\circ$. While the reason for this is unclear, it is plausible that the CNNs have learnt a preference for horizontal and vertical lines. This is intuitive as natural ``unrotated'' images tend to have a horizontal ground plane and vertical buildings or trees which can cause it to learn such preferences. While this seems to explain the curves in Figure \ref{fig:abs} and \ref{fig:hist}, it does not explain Figure \ref{fig:rot_inv}.

\begin{figure}
	\centering
	\includegraphics[width = 9cm,trim={1cm 0cm 0cm 1cm},clip]{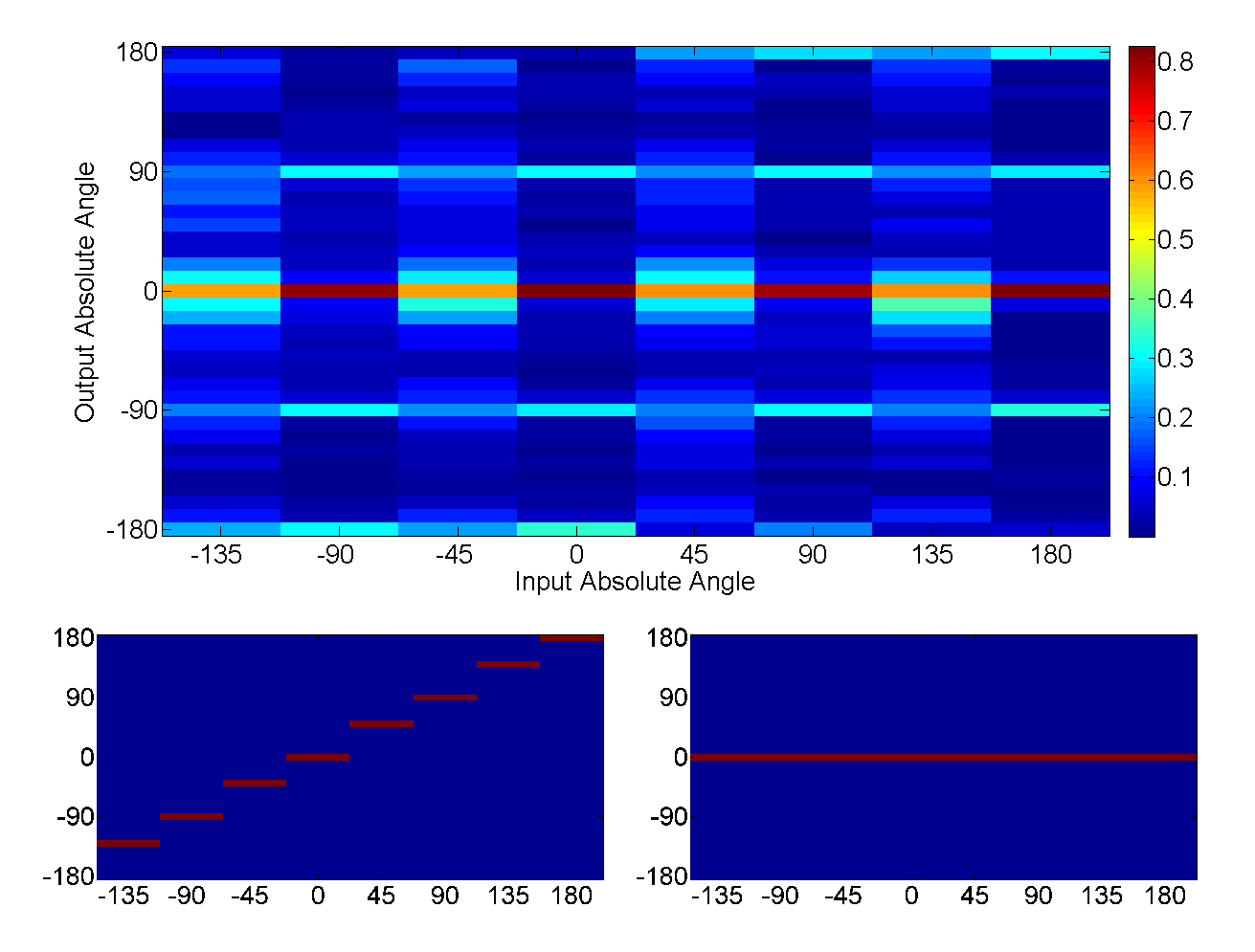}
	\vspace{-0.25cm}
	\caption{{A colour-map visualization of the input-output characteristics of our method. Brighter colour represents higher likelihood of occurance of that input-output pair. The plot on top shows the performance of 3T-CNN. The bottom-left plot is the baseline performance where rotated images remain uncorrected. The figure on bottom-right is the ideal case where all images are corrected to have an absolute angle of $0^\circ$.}} 
\label{fig:imagesc}
\end{figure}

To further investigate the behaviour of our method, we plot the input-output likelihood in Figure \ref{fig:imagesc}. Bright spots indicate high likelihood of an input absolute angle (x-axis) being compensated to a corresponding absolute angle (y-axis). As an example, for an input angle of $-45^\circ$, the most probable output absolute angle is $0$ (which is expected). The figure shows that our method performs well for input absolute angles of $0^\circ$,$~\pm 90^\circ$ and $180^\circ$, as compared to other angles. For other input angles, output is a smooth symmetric distribution centered at $0^\circ$. For those input angles where the method shows high accuracy ($0^\circ$,$~\pm 90^\circ$ and $180^\circ$), it is observed that the images which are not corrected to an absolute $0^\circ$ are mostly likely to be rotated to$~\pm 90^\circ$ or $~\pm 180^\circ$. This behaviour is indeed puzzling, and points to a possibly peculiar property of natural images. 

\begin{figure}
\centering
\includegraphics[width = 8cm,trim={0cm 0cm 0cm 0cm},clip]{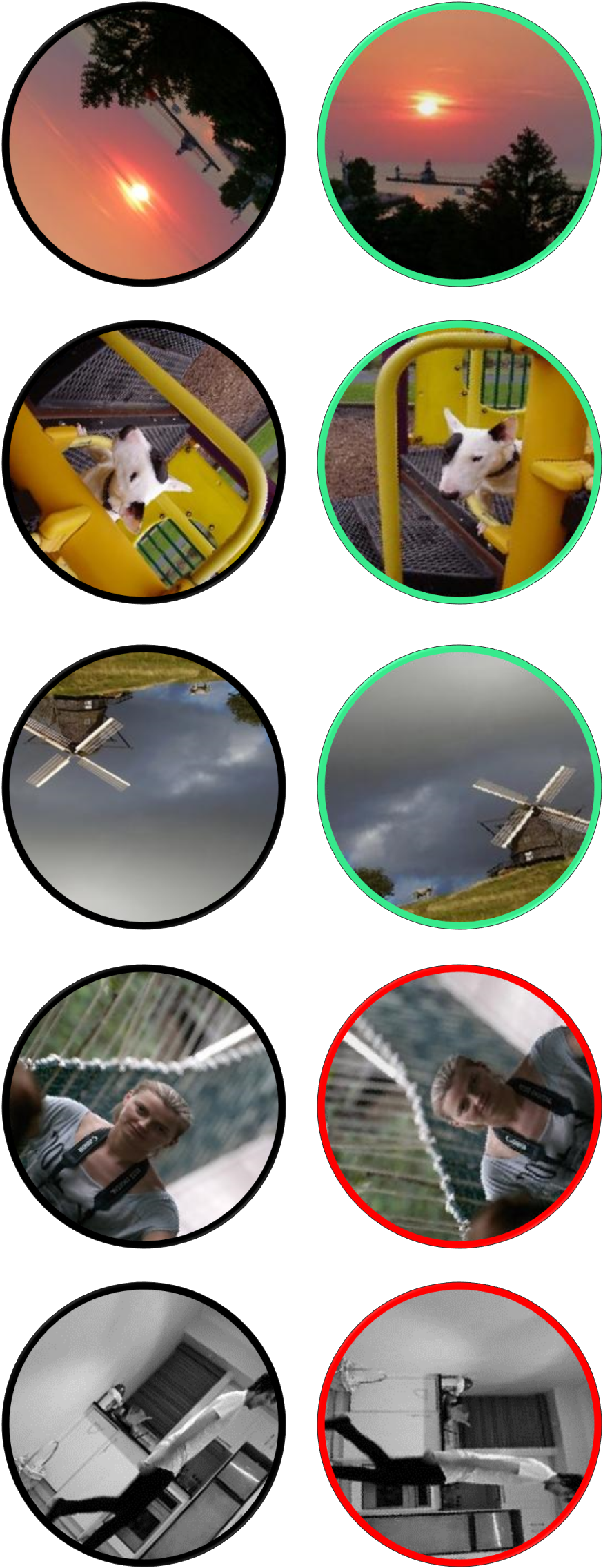}
\caption{Sample input-output pairs of our system. The green circles represent correct compensations, while red circles represent incorrect ones. Despite the presence of faces and people, our model can go wrong.}
\label{fig:im}
\end{figure}

\section{Concluding remarks}
We have presented a method to compensate for large in-plane rotations in natural images. We saw that this is a problem which is much harder than it appears. While our method seems to work well empirically, we noticed certain trends in the kind of mistakes it made. The method seems to make more mistakes for indoor images than it does for outdoor images. It makes such mistakes despite of the presence of humans in the scene. We hypothesize that the lack of a face/person detector in a network as shallow as ours could lead to this. Having said that, our framework is quite general and can be used as a pre-processing step in general computer vision applications to make them more robust to rotations. 

\bibliographystyle{abbrv}
\bibliography{sigproc}
\end{document}